\documentclass[Afour,sageh,times]{sagej}

\usepackage{moreverb,url}
\usepackage{enumerate}
\usepackage{enumitem}
\usepackage{amsmath}
\usepackage{tikz}
\usepackage[colorlinks,bookmarksopen,bookmarksnumbered,citecolor=red,urlcolor=red]{hyperref}
\usepackage{float}
\usepackage{subfig}
\usepackage{gensymb}
\usepackage{verbatim}
\usepackage{comment}
\usepackage{dblfloatfix}
\usepackage{algorithm}
\usepackage[noend]{algpseudocode}
\DeclareMathOperator*{\argmin}{arg\,min}

\newcommand\BibTeX{{\rmfamily B\kern-.05em \textsc{i\kern-.025em b}\kern-.08em
T\kern-.1667em\lower.7ex\hbox{E}\kern-.125emX}}

\def\volumeyear{2019}
\newcommand{\pcite}[1]{(\cite{#1})}

\begin{document}

\runninghead{Serlin et. al.}

\title{Distributed and Consistent Multi-Image Feature Matching via QuickMatch}

\author{Zachary Serlin\affilnum{1}, Guang Yang\affilnum{2}, Brandon Sookraj\affilnum{1}, Calin Belta\affilnum{1}, and Roberto Tron\affilnum{1}}

\affiliation{\affilnum{1}Boston University, Department of Mechanical Engineering, USA \\
\affilnum{2} Boston University, Division of Systems Engineering, USA}

\corrauth{Zachary Serlin, Boston University,
Department of Mechanical Engineering,
110 Cummington Mall,
Boston, Massachusetts, USA.}

\email{zserlin@bu.edu}

\begin{abstract}
In this work we consider the multi-image object matching problem, extend a centralized solution of the problem to a distributed solution, and present an experimental application of the centralized solution. Multi-image feature matching is a keystone of many applications, including simultaneous localization and mapping, homography, object detection, and structure from motion. We first review the QuickMatch algorithm for multi-image feature matching. We then present a scheme for distributing sets of features across computational units (agents) that largely preserves feature match quality and minimizes communication between agents (avoiding, in particular, the need of flooding all data to all agents). Finally, we show how QuickMatch performs on an object matching test with low quality images. The centralized QuickMatch algorithm is compared to other standard matching algorithms, while the Distributed QuickMatch algorithm is compared to the centralized algorithm in terms of preservation of match consistency. The presented experiment shows that QuickMatch matches features across a large number of images and features in larger numbers and more accurately than standard techniques.
\end{abstract}

\keywords{Computer vision, Feature matching, Object matching, Distributed matching, Multi-image matching}

\maketitle

\section{Introduction}
\subsection{Motivation}

Matching is a fundamental operation common in robotics and computer vision algorithms such as Structure from Motion (SfM), Simultaneous Localization And Mapping (SLAM), and object detection and tracking. Matching in these domains is typically performed on feature descriptors (such as SIFT, ORB, and SURF by \cite{SIFT}, \cite{ORB}, and \cite{SURF} respectively) which extract discriminative characteristics from high dimensional data (i.e. an image). At a fundamental level, these features are used to associated unique objects in the universe to their appearance in multiple views. These different views might be acquired by a geographically dispersed camera or robotic network, which can produce large amounts of data. It is therefore important to have the ability to consistent match features across multiple images (i.e, perform multi-image matching) efficiently, consistently, accurately, and, ideally, in a distributed manner. 
For instance, match quality is key in settings such as SfM and SLAM, because the quality of the reconstruction is a direct result of how well the features across images match. The speed of image matching is extremely critical in application such as real-time object detection, since the algorithm must handle many images a second. Cross image consistency -- matching the same object feature across many images consistently -- is also critical for SfM and SLAM since feature points must be tracked not just between images, but across multi-image sequences for the best results. The pervasiveness of graphical processing units (GPUs), networked systems, distributed robotics, cloud computing, and multi-core processors, has made the distributability of computer vision solutions important to their wide applicability. These tools have also shown great advantages in speed and bandwidth for other classic algorithms \pcite{gpu_BF,acel_sift}. To this end, we also introduce a distributed version of our solution to distribute the computational load of the feature matching problem over multiple agents; this allows the algorithm to manage considerably more features, with only a slight performance loss, by spreading the required computations across the network of agents while minimizing the amount of communication. %Note that similar benefits would arise when applying this approach to the setting of single agents with multiple parallel computational units.       

\subsection{Problem overview and contributions}
We propose a solution to the following problem: given a set
of images taken from a team of robots (or camera network), match unique
object features in a distributed manner, as they enter and exit the images
from multiple perspectives. This problem arises often in object
detection, localization, and tracking \pcite{opencv,cunningham2012,Zhou_quickmatch},
homography estimation \pcite{CVAlgo}, structure from motion
\pcite{homography1}, and formation control \pcite{visionformation}.
Solutions to this problem are traditionally computationally complex, and
often mismatch features when considering more than two images 
\pcite{opencv,SIFT}, by either missing matches between two or more views of the same entity in the universe, or by introducing associations between separate universe entities. Multi-image correspondences allow for greater match
reliability, and a more accurate representation of objects in the
universe. The proposed solution leverages a relatively recent algorithm,
QuickMatch \pcite{tron_quickmatch}, to quickly and reliably discover
correspondences across multiple images. The experiments presented in this
paper benchmark QuickMatch's performance by implementing an object
matching framework under realistic conditions (i.e. images with clutter,
repeated structures, and poor image quality); a target object is matched
across a network of cameras, and then these matches are used to generate
the target's trajectory. The solution presented also extends QuickMatch to a distributed computation setting that largely preserves match quality while minimizing communication across agents where possible.

\subsection{Related Work}
Feature matching is a basic process in many computer vision algorithms. \textit{Pairwise matching} is the classical approach to this task, where feature descriptors between two images are compared based on a distance metric (e.g. Euclidean or Manhattan distance), and declared a match if this distance is below some threshold \pcite{SIFT,VLFeat}. Pairwise matching is often posed as a nearest neighbor search problem \pcite{gpu_BF,forests}. This method is used in two standard algorithms, \textit{Brute Force} (BF) matching \pcite{opencv}, and \textit{Fast Library for Approximate Nearest Neighbors} (FLANN) matching \pcite{Dist_NN}. Pairwise matching has difficulties consistently matching entities with repetitive structure or similar appearance (e.g. windows) because the distance metric alone does not consider the \textit{distinctiveness} (smallest distance between features from the same image) of the features. Including distinctiveness of features during matching has been shown to be beneficial \pcite{SIFT}. For multi-image matching, pairwise matches scale poorly with the number of images and across multiple images, match correspondences often do not belong to the same ground truth object (and hence link together correspondences between different entities). \textit{Graph matching} has also been used for pairwise matching. This approach attempts to match vertices (features) and edges (matches) simultaneously to determine better pairwise matches shown by \cite{multi_graph} and \cite{multi_graph_unified}, but it cannot handle the multi-image setting for the same reasons as above. 

Pairwise matching can also be approached as an unsupervised learning problem, and therefore there are a number of distributed approaches to the classic algorithms, especially for the k-nearest neighbor problem. The k-nearest neighbor problem has been extensively explored in settings such as friend suggestion on Facebook, image classification, and recommendation systems. For this reason, many parallelized approaches have been introduced by \cite{billion_knn}, \cite{acel_nn}, and \cite{buffer_kd}. In a feature matching setting, a CUDA based distributed approach to the BF algorithm has been proposed that is approximately 100x faster than its centralized counterpart. Beyond matching, parallel computing has also been applied to feature extraction algorithms such as SIFT \pcite{acel_sift}, offering speed increases along the entire feature matching pipeline. These approaches still do not consider the multi-image matching problem, however they offer a strong motivation for distributing existing multi-image matching techniques. Machine learning techniques have also been used to extract more meaningful features that are characteristic of objects in general, which adds a layer of semantic understanding to the multi-image matching problem \pcite{hypercolumns,wang2018multi,anchornet}.   

Beyond pairwise matching, a number of other approaches exist for multi-image matching (where multiple images are directly considered) that are based on optimization, graphs, and clustering. \textit{Optimization} based approaches are based upon non-convex problems where optimization constraints must often be relaxed to reliably obtain solutions \pcite{optimal_point_corespond,multi_graph,chen_icml14,data_ass2017,multi_perm}. Moreover, these approaches typically require \textit{a priori} the number of objects, which is often not available, and they do not consider distinctiveness of the features. \textit{Cycles in graphs} are early predecessors to the QuickMatch algorithm and have largely been used to remove inconsistent matches as shown by \cite{cycles}. Cycle consistency has also recently been used in a distributed manner to perform matching by \cite{Dis_multi_image2018} and \cite{consistent_2011}. \textit{Clustering} can be cast as finding clusters of similar features. Algorithms such as k-means \pcite{info_theory} and spectral clustering \pcite{Spectral_clustering} have been explored to this end, but also often require a predefined number of objects, and do not consider that a unique feature only occurs once in an image, meaning repeat structures are also often called a match.   

QuickMatch is grounded in \textit{density-based clustering} algorithms (see work by \cite{gradient_density,quickshift,DBSCAN,psdbscan} for examples), which find clusters by estimating a non-parametric density distribution of data \pcite{pdf,density_function}. These approaches do not require prior knowledge of the number or shape of clusters, and can be modified to include feature distinctiveness by construction. 

This article is an extension of previous work presented at the International Symposium of Experimental Robotics \cite{iser2018} and the International Conference on Computer Vision \cite{tron_quickmatch}. The primary extensions from this prior work are:
\begin{itemize}[leftmargin=*,noitemsep]
    \item We introduce a method for distributing the matching problem across multiple agents based on a feature space partition.
    \item We demonstrate the distributed solution is nearly equivalent to the centralized feature matching solution and minimizes inter-agent communication. 
\end{itemize}

The primary contributions of this paper are firstly the testing and experimental validation of the QuickMatch algorithm under more realistic conditions, as opposed to previous evaluations using standard data sets. This experiment tests the algorithm for computational efficiency and match accuracy by employing a distributed camera network to localize a moving target. Secondly, this paper introduces the Distributed QuickMatch (DQM) algorithm. This algorithm extends the QuickMatch framework to handle distributed computation with minimal loss to match quality, and minimal inter-agent communication.

\subsection{Outline}
The remainder of this paper is structured as follows. We begin by introducing preliminary concepts for features, feature space partitions, and multi-image matching. We then present the multi-image matching problem. Next, we present first the centralized, and then decentralize solutions to the presented problem, and we detail the theory of the approach preformed in the experiment. We then present simulations to compare the centralized and distributed solutions. Finally we present the experiment preformed and its results.

\section{Preliminaries}\label{sec:prelims}
\subsection{Images and Features}
We start by considering a set of images $\mathcal{I} = \{1,2, \ldots, i, \ldots, N\}$ where $i \in \mathcal{I}$ denotes a single image. From each image $i$, we extract a set of features $K_i$ with a single feature vector denoted $x_{ik} \in \mathbb{R}^F$, where $F$ is the dimension of the feature space (eg. for SIFT, $F=128$), $i$ is the image index, and $k$ is the feature index in that image. We denote the set of all features as $K_\mathcal{I}$ and the cardinality of a set as $\vert K_\mathcal{I} \vert$.

\subsection{Agents and Feature Space Partitions}
In the distributed computational setting, we consider a set of computational units, or agents, $A=\{1,2, \ldots, a, \ldots, m\}$ where $a \in \mathcal{A}$ denotes a single agent. The feature space is denoted as $\mathcal{Y} = \mathbb{R}^F$, and is partitioned into $m$ convex Voronoi tessellation subspaces, where each space is assigned to a single agent as $\mathcal{Y}_a \subseteq \mathcal{Y}$ with $\bigcup\limits_{a \in A}\mathcal{Y}_a = \mathcal{Y}$. The Voronoi partition seed of each $\mathcal{Y}_a$ is given as $P_a \in \mathbb{R}^F$. We define a labeling function $\ell:x_{ik} \mapsto a$ that maps a feature $x_{ik}$ to a given agent depending on which $\mathcal{Y}_a$ it exists in. This labeling function also has a set-valued inverse, $\ell^{-1}:a \mapsto x_{ik}$, which returns the set of features contained in any given agent partition.

\subsection{Multi-Image Matching}
The multi-image matching problem presented here is summarized from \cite{tron_quickmatch} and is posed in the light of both cycle consistency in a graph of features, and clustering of features. We refer the reader to the original paper for detailed proofs regarding the equivalence between the two views of the problem.

We begin by assuming that $\mathbb{R}^F$ has a distance metric in the feature space defined as $d(x_{ik},x_{i'k'}) \mapsto \mathbb{R} \geq 0$ between two features. We denote the set of all features $\mathcal{X} = \{x_{ik}\}^{i \in \mathcal{I}}_{k \in 1 \ldots N}$. When two features are a match, i.e., $x_{i_1k_1} \rightarrow x_{i_2k_2}$, $i_1 \neq i_2$, we mean that the two features represent the same entity. Based on this type of correspondence, we define a directed graph of the matches as $\mathcal{G} = (\mathcal{V},\mathcal{E})$ with $\mathcal{V} = \{x_{ik}\}$ and $\mathcal{E} \subset \mathcal{V} \times \mathcal{V}$ is the set of matches $x_{i_1k_1} \rightarrow x_{i_2k_2} \in \mathcal{E}$.

We define multi-image matches as subsets of the features in $\mathcal{X}$. We therefore define, given a subset of features $\mathcal{C} \subset \mathcal{V}$, a subgraph of $\mathcal{G}$ restricted to $\mathcal{C}$ as $\mathcal{G} \vert_\mathcal{C} = (\mathcal{C},\mathcal{E}')$ with $\mathcal{E}' = \{x_{i_1k_1} \rightarrow x_{i_2k_2} \in \mathcal{E}:x_{i_1k_1},x_{i_2k_2} \in \mathcal{C}\}$. A graph $\mathcal{G}=(\mathcal{V},\mathcal{E})$ is connected if for any $x_{i_1k_1},x_{i_nk_n} \in \mathcal{V}$ there exist a sequence $x_{i_1k_1},x_{i_2k_2},\ldots,x_{i_nk_n}$, called a path, such that $x_{i_j,k_j} \rightarrow x_{i_{j+1},k_{j+1}} \in \mathcal{E} \forall j \in \{1,\ldots,n-1\}$ and a cycle is a path where the start and end features are then same, or $x_{i_1k_1} = x_{i_nk_n}$. We define a clique as a graph $\mathcal{G}$ with $x_{i_1k_1} \rightarrow x_{i_2k_2} \in \mathcal{E} \forall x_{i_1k_1},x_{i_nk_n} \in \mathcal{V}$. 

A set of multi-image matches is a set $\mathcal{M}=\{\mathcal{C}_c\}$ of clusters $\mathcal{C}_c = \{x_{i_1k_1},\ldots,x_{i_nk_n}\}$, where each one corresponds to a single entity in the universe, such that
\begin{enumerate}
    \item [(C1)] $\mathcal{M}$ is a partition of $\mathcal{X}$ (i.e., each feature $x_{ik}$ appears exactly in one set $\mathcal{C}_c$);
    \item [(C2)] Each set $\mathcal{C}_c$ has at most one feature per image;
    \item [(C3)] There is an induced directed graph $\mathcal{G}_\mathcal{M} = (\mathcal{X},\mathcal{E}_\mathcal{M})$ of pairwise mathces such that, for any $\mathcal{C}_c \in \mathcal{M}$, the subgraph $\mathcal{G}_{\mathcal{M}} \vert \mathcal{C}_c$ is a clique (i.e., $\mathcal{G}_\mathcal{M}$ is a union of cliques).
\end{enumerate}

Conditions (C1) and (C2) ensure that the same entity cannot appear in multiple clusters, and an entity cannot appear more than once per image. Condition (C3) requires that features from the same cluster always match.

Given the above statements, three properties are implied about $\mathcal{G}_\mathcal{M}$,
\begin{enumerate}
    \item [(P1)]  Symmetry: $x_{i_1k_1} \rightarrow x_{i_2k_2} \in \mathcal{E}_\mathcal{M}$ implies $x_{i_2k_2} \rightarrow x_{i_1k_1} \in \mathcal{E}_\mathcal{M}$; 
    \item [(P2)] Cycle Constraint: Given a path $x_{i_1k_1},\ldots,x_{i_nk_n}$ in $\mathcal{G}_\mathcal{M}$, having $i_1 = i_n$ implies $k_1=k_n$ meaning the path is a cycle;
    \item [(P3)] Single match: A feature cannot correspond to two different features in another image meaning, if $x_{i_1k_1} \rightarrow x_{i_2k_2}$ and $x_{i_1k_1} \rightarrow x_{i_2k'_2}$ belong to $\mathcal{G}_\mathcal{M}$ then $k_2 = k'_2$.
\end{enumerate}

\section{Problem Statement}
Given a set of images $\mathcal{I} = \{1,2, \ldots, i, \ldots, N\}$ and a set of $K_i$ feature vectors, $x_{ik}$, extracted from each image, determine matches ($x_{i_1k_1}\leftrightarrow x_{i_2k_2} : i_1 \neq i_2$) between features from separate images, such that matched features represent the same point in the scene and build a set of multi-image matches $\mathcal{M}$.

\section{Solution}
We present both a centralized and decentralized solution to this problem. It is important to note that this problem cannot be solved with only pairwise matches as they are not capable of considering cliques directly; specifically, they cannot prevent the violation of (C3). The centralized solution introduces the QuickMatch algorithm and demonstrates its effectiveness with a two-stage, offline, centralized implementation on a system of distributed ground robots and a central computer. Features are first extracted using off-the-shelf feature extraction methods (SIFT), and the features are then matched using the QuickMatch algorithm to find a given reference object. These matches are used to perform homography estimation between the reference object and the camera network to generate target trajectories. The decentralized solution then builds on the QuickMatch algorithm with a framework for distributing the features across multiple computational units while maintaining match consistency. This framework can be used for other matching algorithms as well, however we compare its results with the centralized solution to demonstrate near equivalence to the centralized solution.

\subsection{Feature Extraction}
Feature extraction aims to find and describe representative points from high dimensional data, such as an image \pcite{SURF,ORB,SIFT,Zhou_quickmatch}. Features themselves are also high dimensional vectors but typically of much lower dimension then the original data. In this experiment, the \textit{scale invariant feature transform} (SIFT) feature is used, which extracts $K_i$ 128-dimensional vectors that represent the appearance of each feature point. See \cite{SIFT}, and \cite{VLFeat} for more details on this standard feature extraction algorithm. Other feature types can be used and we also tested with Oriented FAST and Rotated BRIEF (ORB) features and Speeded-Up Robust Features (SURF), however SIFT was qualitatively the most reliable. 
\subsection{QuickMatch}
The QuickMatch algorithm begins by calculating the distance between all features (we use Euclidean distance). For each image, the minimum distance, $\sigma_i$, between any two features is used as the distinctiveness of features for that image. Recall, from above, $x_{ik}$ is a point in the high dimensional feature space. The feature density $D(x_{ik})$ is then calculated for each point using the formula
\begin{eqnarray}\label{eq:density}
D(x) = \sum_{i=1}^N \sum_{k=1}^{\vert K_i \vert} h(x,x_{ik};\sigma_i),\\
h(x_1,x_2;\sigma_i) = \exp(-\frac{\Vert x_1 - x_2 \Vert}{2\sigma_i^2}),\\
\sigma_i = \min\limits_{\forall k,i=i'}d(x_{ik},x_{i'k'}),
\end{eqnarray}
with kernel function $h$, and distinctiveness $\sigma_i$. With this feature density, the features are organized into a tree structure, with parent nodes being the nearest neighbor with a higher density, 
\begin{eqnarray}
T: parent(x_{ik}) = \min\limits_{i^\prime k^\prime \in J} d(x_{ik},x_{i^\prime k^\prime}),\\
J = \{ i^\prime k^\prime : k \neq k^\prime, D(x_{i^\prime k^\prime})>D(x_{ik})\}.
\end{eqnarray}
Edges are directed to parents along the gradient of feature density, and ultimately toward the center of the parent cluster or to another distant cluster. Once the tree has been constructed, edges are broken based on (C1)-(C3) above if either of two criteria are met;
\begin{enumerate}
    \item If parent and child groups have nodes from the same image (i.e. $i_1 = i_2$).
    \item If the edge is larger than a user defined threshold ($\rho$) times the distinctiveness $\sigma_i$ (i.e. $d(x_{ik},parent(x_{ik})) \geq \rho \sigma_i$).
\end{enumerate}
This method results in a forest of trees ($\mathcal{M}$), where each tree is a cluster ($C_c$) representing a  unique entity in the universe. In practice, each tree represents a point that is common among images, meaning the algorithm discovers common features among very similar objects. Feature discovery will be explored further in the results section, where groups of matching points are employed for object detection and homography transformations. 
%\vspace{-4pt}
\begin{algorithm} 
\caption{QuickMatch}\label{algo:QM}
\begin{algorithmic}
    \State Input: $K,\rho$
    \State Output: Clusters $\mathcal{C}_c$ ($\mathcal{M}$)\\
    %\dotfill
    \ForAll{$x_{ik},x_{i'k'}$}
        \State Compute $h(x_{ik},x_{i'k'},\sigma)$
    \EndFor
    \ForAll{$x_{ik}$}
        \State Compute $D(x_{ik})$
    \EndFor
    \ForAll{$x_{ik}$}
        \State Compute $parent(x_{ik})$ to build $\mathcal{T}$
    \EndFor
    \For{edges in $T$} 
    \Comment{From shortest to longest}
        \State $x_{ik} \in \mathcal{C}_c,x_{i'k'} \in \mathcal{C}_{c'}$ are ends of edge
        \If{$\{i\} \in \mathcal{C}_c \cap \{i\} \in \mathcal{C}_{c'} = \emptyset$ \textbf{and} $d(x_{ik},x_{i'k'})\leq \rho \argmin\limits_{\sigma}({\mathcal{C}_{c},\mathcal{C}_{c'}})$}
            \State Merge $\mathcal{C}_{c},\mathcal{C}_{c'}$
        \Else
            \State Remove edge
        \EndIf
    \EndFor
    \Return $\mathcal{M}$
\end{algorithmic}
\end{algorithm}

\subsection{Distributed QuickMatch}
At a high level, the distributed version of QuickMatch follows two rounds:
\begin{enumerate}
    \item Each node sends each feature to a different node \emph{depending on its location} (i.e., each node receives all the features belonging to its own element of a partition of the feature space); each node performs QuickMatch independently.
    \item Nodes identify clusters falling near the edges of the partition; if necessary, these clusters are transferred to other nodes for re-processing.
\end{enumerate}
At a high level, each node needs to process only a subset of the entire dataset, and, more importantly, each point is typically transferred only once, without having to flood the entire network with the entire dataset (additional transmissions are required only for edge cases, which are typically much rarer).
Details of each operation in Distributed QuickMatch are presented below.

\subsubsection{Feature Space Partitioning}
The goal of partitioning the feature space is to split the computation load of QuickMatch among $m$ agents, ultimately allowing for more features to be matched, while maintaining multi-image match quality. To do this, instead of naively partitioning the space with an algorithm like k-means and sending features to their nearest Voronoi partition, we build trees of features that can later be broken according to the QuickMatch rules.   

We modify the labeling function presented in the preliminary section to denote round being considered as $\ell(x_{ik},r) \mapsto a$ and $\ell^{-1}(a,r) \mapsto \{x_{ik}\}$. 
In practice this round is $r=\{0,1\}$, where $0$ is the initial partitioning, and $1$ is the labeling after partition reassignment. The algorithm begins with a naive partitioning of the features
\begin{equation}\label{eq:init}
    \ell(x_{ik},0) = \argmin\limits_{a \in A}d(x_{ik},P_a) \forall x_{ik}\in K_\mathcal{I},
\end{equation}
where the $P_a$ Voronoi seeds are found using the k-means algorithm \cite{k-means}. More explicitly, we choose random points in the feature space $P_a^{(0)}, \forall a\in A$ and build an initial partition as 
\begin{multline}\label{eq:kmeans1}
    \mathcal{Y}_a^{(t)} = \{x_{ik} : \Vert x_{ik}-P_a^{t} \Vert^2 \leq \Vert x_{ik}-P_{a'}^{t} \Vert^2 \\ \forall a'\in A, a' \neq a,
\end{multline}
with each $x_{ik}$ assigned to only one agent and $t$ is the iteration index. The points are then iteratively updated until $t = 100$ iterations according to first equation \ref{eq:kmeans1}, and then 
\begin{equation}\label{eq:kmeans2}
    P_a^{t+1} = \frac{1}{\mathcal{Y}_a^{(t)}} \sum_{x_{ik}\in \mathcal{Y}_a^{(t)}}x_{ik}.
\end{equation}
We denote the set of all Voronoi seeds as $P_A$. In larger data sets, and in a more distributed setting, the choice of $P_A$ can be done at random. By agreeing on a random seed to generate these values from, the agents need to communicate only a single integer. Qualitatively, we ran the random seed procedure on the data sets below and they showed similar matching performance, but with a larger variance in the number of features in each partition.  

\subsubsection{Tree Building}
Once $P_A$ is determined with k-means, we send features to their respective agents based on $\ell(x_{ik},0)$, and at each agent, in parallel, we calculate the density of each feature as in Equation \ref{eq:density}. Unlike QuickMatch however, the density kernel is finite in order to limit interaction of density across the partition boundaries. The density kernel here is now defined as 
\begin{multline}\label{eq:kern2}
    h(x_1,x_2;\sigma) = \begin{cases}
        \Vert x_1 - x_2 \Vert < \sigma & -\frac{(\Vert x_1 - x_2 \Vert)^2}{\sigma} + 1, \\
        \text{else} & 0,
    \end{cases}
\end{multline}
which is a finite quadratic kernel. 

Once the density is calculated with the kernel above, we build a tree from the features in each agent's partition according to
\begin{eqnarray}\label{eq:treea}
    T_a: parent(x_{ik}) = \min\limits_{i^\prime k^\prime \in J} d(x_{ik},x_{i^\prime k^\prime}),\\
J = \{ i^\prime k^\prime : k \neq k^\prime, D(x_{i^\prime k^\prime})>D(x_{ik}), x_{ik}\in \mathcal{Y}_a\}. 
\end{eqnarray}

Similarly to the distinctiveness in QuickMatch, we now introduce a distinctiveness metric that is agent specific and given as 
\begin{equation}
    \sigma_a = \max{(\sigma_p)} \forall i \in \ell^{-1}(a,0),
\end{equation}
where
\begin{equation}
    \sigma_p = d(x_{ika},parent(x_{ika})),
\end{equation}
where $x_{ika}$ represents feature $x_{ik} \in \mathcal{Y}_a$.

With the feature space partitioned by the above seeds $P_A$, we now determine if individual features may belong to clusters split by these artificial boundaries. To do this, we must determine the distances between the features and these boundaries before we can reason about split cluster membership.

\subsubsection{Partition Boundary Distance}
To find the minimum distance between a feature and the boundaries that are generated by the Voronoi partition, we formulate the following problem: Consider a set of root nodes $\{P_0,...,P_m\}$ and their corresponding regions $\{\mathcal{Y}_0,...,\mathcal{Y}_m\}$, given a test node $x_t \in \mathcal{Y}_t$ and evaluation node $P_e \in \mathcal{Y}_e$, where $\mathcal{Y}_e \subset \mathcal{Y} \setminus \mathcal{Y}_t$, find the minimum distance $d_{min}$, and the corresponding hyper-plane, to $x_t$. A graphical illustration of this problem can be found below in Figure \ref{fig:min_problem_setup}.

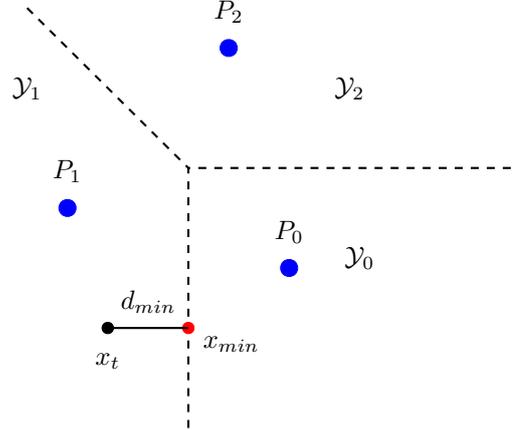
\begin{figure}[h!]
  \begin{center}
    \begin{tikzpicture}[rotate=45,scale=.75]
      \draw [blue,fill] (0,-2.5) circle [radius=0.15] node [black,above=6] {$P_0$}; % Draws P0
      \draw [blue,fill] (-2,1) circle [radius=0.15] node [black,above=6] {$P_1$}; % Draws P1
      \draw [blue,fill] (2,1) circle [radius=0.15] node [black,above=6] {$P_2$}; % Draws P2
      \draw [thick,dashed] (-3.25,-3.25) % Draws a line
      %to [out=10,in=190] (1,3)
      to  (0,0);  
      \draw [thick,dashed] (4,-4) % Draws a line
      %to [out=10,in=190] (1,3)
      to  (0,0);   
      \draw [thick,dashed] (0,4) % Draws a line
      %to [out=10,in=190] (1,3)
      to  (0,0);   
      \draw [fill] (-3,-1) circle [radius=0.1] node [black,below=6] {$x_{t}$}; % Draws x_t
      \draw [red,fill] (-2,-2) circle [radius=0.1] node [black,below=6,right=2] {$x_{min}$}; % Draws x_min
      \draw  [thick] (-3,-1)
      to (-2,-2) node [black,left=15,above=2] {$d_{min}$};
      \draw (1,-3.25) node [black] {$\mathcal{Y}_0$};
      \draw (-1,3) node [black] {$\mathcal{Y}_1$};
      \draw (3,-1) node [black] {$\mathcal{Y}_2$};
    \end{tikzpicture}
    \caption{Illustration of Voronoi boundary distance problem.}
    \label{fig:min_problem_setup}
  \end{center}
\end{figure}

Given the test feature node $x_t$, its current root node $P_t$ and the root node that we want to evaluate $P_e$, we formulate a \emph{Quadratic Program} (QP) as the following:
\begin{equation} 
\begin{aligned}
& \underset{x\in \mathbb{R}^F}{\min}
& & (x_t-x)^T(x_t-x)\\
& \text{s.t.}
& & (\frac{P_e-P_t}{\|P_e-P_t\|})^T (x-P_t) - \frac{d(P_t, P_e)}{2}\geq 0, \\
\end{aligned}
\label{eq:min_dist_QP}
\end{equation}
where $d(P_t, P_e)$ is the pairwise distance between $P_e$ and $P_t$. We obtain the closet point $x_{min}$ on the Voronoi boundary that has the minimum distance to $x_{t}$. By iterating through all possible root nodes $P_e, \forall e \neq t$, we find the global minimum distance and its corresponding root node and its partition index
\begin{equation}\label{eq:dika}
    d_{ika} = d(x_{ika},x_{min}).
\end{equation}

\subsubsection{Contested Feature Re-assignment}
Once we are able to determine the distance between a feature and the partition boundary, we can then determine the contested features in each region. These features are the ones that may require reassignment to another partition's tree. These are the features in danger of being mismatched due to the feature space partitioning splitting their cluster. This contested set of features $S_a$ is defined as 
\begin{multline}\label{eq:ineq_cont}
    S_{a} = \{ a' \in A\setminus a:(d_{ika}+d_{aa'})<\sigma_p\} \\
    \forall x_{i,k} \in \ell^{-1}(a,r) \forall a \in A,
\end{multline}
where
\begin{eqnarray}\label{eq:daa}
    d_{aa'}=\min\limits_{x_{ik}\in\ell^{-1}(a',0)}d_{ika'} \\
    \begin{split}
        d_{ika'} = d(x_{ika'},\mathcal{Y}_{a}) \\
        a'\neq \ell(x_{ik},0) 
    \end{split}
\end{eqnarray}
where $d_{ika}$ is the distance between each feature $x_{ika}$ and the boundary of the partition of $a'$, and $d_{aa'}$ is the minimum distance between the feature in $a'$ and the $\mathcal{Y}_a$ boundary. 

If the inequality in Equation \ref{eq:ineq_cont} is true, there could be a feature in $a'$ that could belong to the same cluster as $x_{ik}$ in the worst case. This test is overly conservative because it considers the worst case, but it also requires a single value $d_{aa'}$ be communicated between each agent to perform the entire check. Figure 2 illustrates a case of this check. In this case, $x_{ika}$ is not contested with $a'$ because the in inequality in Equation \ref{eq:ineq_cont} is false (i.e. $(d_{ika}+d_{aa'})\geq\sigma_p$).

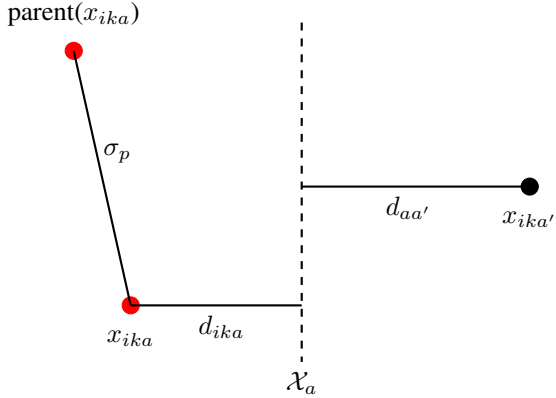
\begin{figure}[h!]
  \begin{center}
    \begin{tikzpicture}[scale=.75]
      \draw [red,fill] (-3,2.5) circle [radius=0.15] node [black,above=6] {parent($x_{ika}$)}; % Draws a rectangle
      \draw [thick,dashed] (1,3) % Draws a line
      %to [out=10,in=190] (1,3)
      to  (1,-3) node [black,below] {$\mathcal{X}_a$};    
      \draw [fill] (5,0.1) circle [radius=0.15] node [black,below=6] {$x_{ika'}$}; % Draws another rectangle
      \draw [red,fill] (-2,-2) circle [radius=0.15] node [black,below=6] {$x_{ika}$}; % Draws a circle
      \draw  [thick] (-2,-2)
      to (1,-2) node [black,left=30,below] {$d_{ika}$};
      \draw  [thick] (5,0.1)
      to (1,0.1) node [black,right=40,below] {$d_{aa'}$};
      \draw  [thick] (-3,2.5)
      to (-2,-2) node [black,left=5,above=50] {$\sigma_p$};
    \end{tikzpicture}
    \caption{Example of comparison to determine if a point is contested.}
  \end{center}
\end{figure}

Once the contested features are determined, we break the agent tree into a forest of trees (clusters) in the same manner as QuickMatch; 
\begin{enumerate}
    \item If parent and child groups have nodes from the same image (i.e. $i_1 = i_2$).
    \item If the edge is larger than a user defined threshold ($\rho$) times $\sigma_i$ (i.e. $d(x_{ik},parent(x_{ik})) \geq \rho \sigma_i$).
\end{enumerate}
Except here, the distinctiveness $\sigma_i$ is calculated for only the features in the agent partition. 

With the clusters formed, each agent then determines which clusters have contested points. Formally, we check $S_a \cap C_c \neq \emptyset$ (i.e. the intersection between a cluster and then contested points is not empty). For each cluster with a contested point we determine the minimum partition index among the contested points, and then send the cluster to that agent now denoted $a'$.

Upon arrival at $a'$, that agent checks if 
\begin{equation}\label{eq:near_contest}
    (\argmin\limits_{x_{ik}\in a'} d(x_{ik},x_{i'k'}) \forall x_{i'k'}\in C_c) \in C_{a'}
\end{equation}
where $C_c$ is the newly acquired cluster. If the nearest point to $C_c \in \mathcal{Y}_{a'}$ is also contested, both $C_c$ and the cluster containing that nearest point from Equation \ref{eq:near_contest} ($C_{c'}$) are sent to the lowest partition index if that index is lower than that agent's index (i.e. clusters are only transferred to agents with lower contested partition index. This lower index requirement prevents switching incomplete clusters repeatedly between agents. This process is performed by decreasing agent index, reducing communication requirements, and the transfer of clusters to at most once per agent.

\begin{algorithm} 
\caption{Distributed QuickMatch}\label{algo:DQM}
\begin{algorithmic}
    \State Input: $K,\rho,\vert A \vert$
    \State Output: Clusters $\mathcal{C}_c$\\
    \dotfill
    \State Compute $P_A$ given $K$
    \Comment Eq. \ref{eq:kmeans1} and \ref{eq:kmeans2}
    \State Compute $\ell(x_{ik},0)$ 
    \Comment Eq. \ref{eq:init}
    \ForAll{$a \in A$}
        \ForAll{$x_{ika},x_{i'k'a}$}
            \State Compute $h(x_{ika},x_{i'k'a},\sigma_i)$
            \Comment Eq. \ref{eq:kern2}
        \EndFor
        \ForAll{$x_{ika}$}
            \State Compute $D(x_{ika})$
            \Comment Eq. \ref{eq:density}
        \EndFor
        \ForAll{$x_{ik}$}
            \State Compute $parent(x_{ika})$ to build $\mathcal{T}_a$
            \Comment Eq. \ref{eq:treea}
        \EndFor
        \ForAll{$x_{ik}$}
            \State Compute $d_{ika}$
            \Comment Eq. \ref{eq:dika}
        \EndFor
        \ForAll{$a \in A$}
            \ForAll{$a' \in A, a' \neq a$}
                \State Compute $d_{aa'}$
                \Comment Eq. \ref{eq:daa}
            \EndFor
        \EndFor
         \ForAll{$x_{ika}$}
            \If{$d_{ijk} + d_{aa'} \leq \sigma_p$}
                \State $x_{ika} \in S_a$
                \Comment Eq. \ref{eq:ineq_cont}
            \EndIf
        \EndFor
        \ForAll{edges in $\mathcal{T}_a$} 
            \If{$\{i\} \in \mathcal{C}_c \cap \{i\} \in \mathcal{C}_{c'} = \emptyset$ \textbf{and} $d(x_{ika},x_{i'k'a})\leq \rho \argmin\limits_{\sigma}({\mathcal{C}_{c},\mathcal{C}_{c'}})$}
                \State Merge $\mathcal{C}_{c},\mathcal{C}_{c'}$
            \Else
                \State Remove edge from $\mathcal{T}_a$
            \EndIf
        \EndFor
        \ForAll{$C_c \in \mathcal{Y}_a$}
            \If{$C_c \cap S_a \neq \emptyset$}
                \State Compute $\min index(a')$ of $C_c \cap S_a$
                \If{$index(a') < index(a)$}
                    \State Send $C_c$ to $a'$
                \EndIf
            \EndIf
        \EndFor
    \EndFor
    \ForAll{$a \in A$}
    \Comment From high to low index
        \If{$C_c$ received from any $a'$}
            \If{$(\argmin\limits_{x_{ika}\in a} d(x_{ik},x_{i'k'}) \forall x_{i'k'}\in C_c) \in S_{a}$} 
                \State Compute $\min index(a')$ of $C_c \cap S_a$
                \If{$index(a') < index(a)$}
                    \State Send $C_c$ to $a'$
                \EndIf
            \EndIf
        \EndIf
    \EndFor
    \ForAll{$a \in A$}
         \ForAll{$x_{ika},x_{i'k'a}$}
            \State Compute $h(x_{ika},x_{i'k'a},\sigma_i)$
            \Comment Eq. \ref{eq:kern2}
        \EndFor
        \ForAll{$x_{ika}$}
            \State Compute $D(x_{ika})$
            \Comment Eq. \ref{eq:density}
        \EndFor
        \ForAll{$x_{ik}$}
            \State Compute $parent(x_{ika})$ to build $\mathcal{T}_a$
            \Comment Eq. \ref{eq:treea}
        \EndFor
        \ForAll{edges in $\mathcal{T}_a$} 
            \If{$\{i\} \in \mathcal{C}_c \cap \{i\} \in \mathcal{C}_{c'} = \emptyset$ \textbf{and} $d(x_{ika},x_{i'k'a})\leq \rho \argmin\limits_{\sigma}({\mathcal{C}_{c},\mathcal{C}_{c'}})$}
                \State Merge $\mathcal{C}_{c},\mathcal{C}_{c'}$
            \Else
                \State Remove edge from $\mathcal{T}_a$
            \EndIf
        \EndFor
    \EndFor
\Return All $C_c (\mathcal{M})$
\end{algorithmic}
\end{algorithm}

\begin{figure*}[!b]
\captionsetup[subfigure]{justification=centering}
\centering
  \subfloat[]{\includegraphics[width=.32\linewidth]{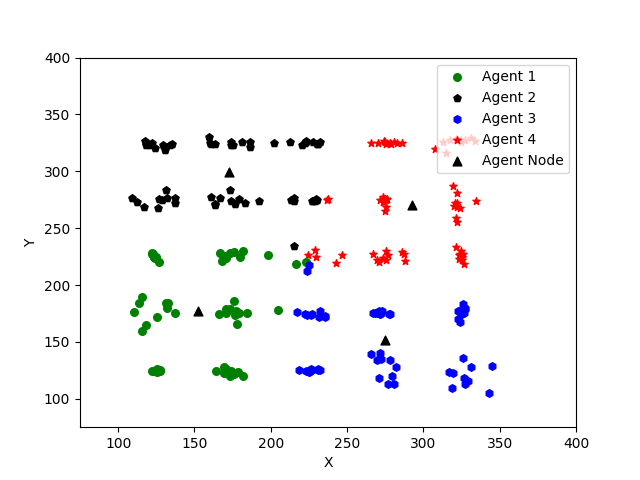}}
  \subfloat[]{\includegraphics[width=.32\linewidth]{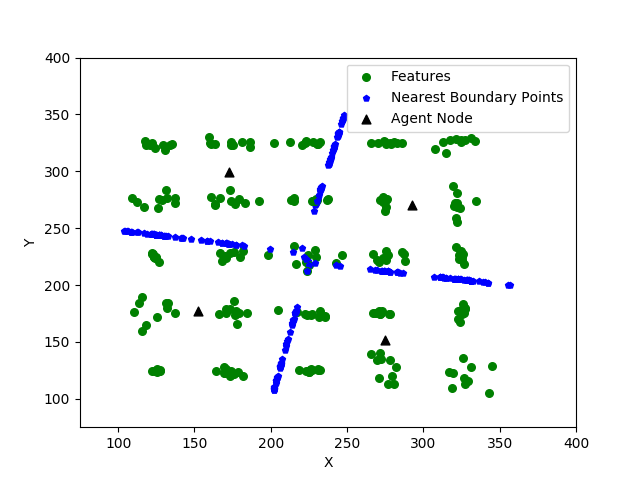}}
  \subfloat[]{\includegraphics[width=.32\linewidth]{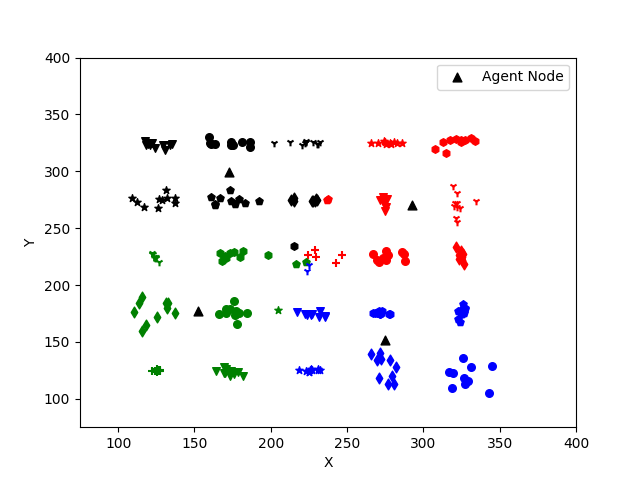}}\\
   \subfloat[]{\includegraphics[width=.4\linewidth]{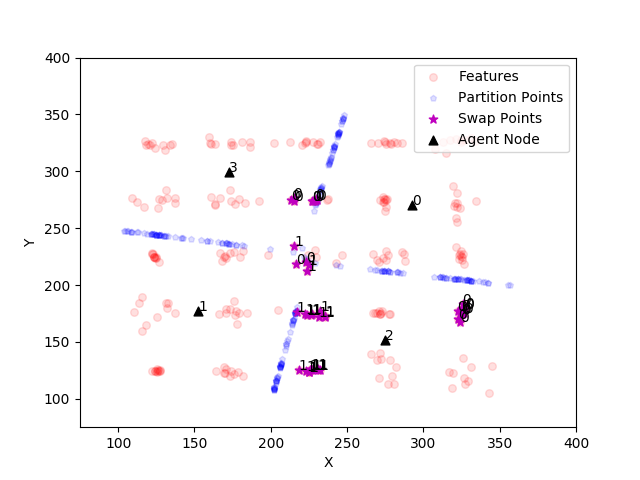}}
   \subfloat[]{\includegraphics[width=.4\linewidth]{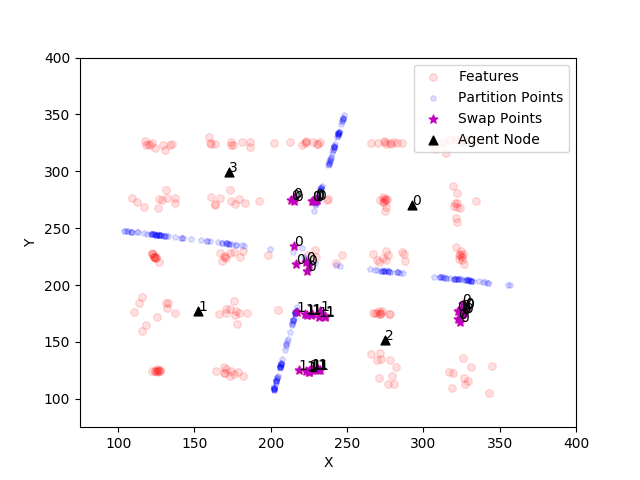}}
  \caption{(a) Synthetic data set features with naive k-means partition assignments. (b) Synthetic data set with $x_{min}$ boundary points plotted to highlight partition. (c) QuickMatch solution for each agent with k-means partition. (d) Initial proposed feature switches of contested clusters. Note the values on the highlighted features denote the proposed $a'$ partition index. (e) Final proposed switch assignments after partitioning section of Distributed QuickMatch is complete.}
  \label{fig:sim1}
\end{figure*}

The final step in reassignment is to rerun the tree building and breaking routines from above to reform all of the clusters. Although this is computationally intensive, it does not require any further inter-agent communication. It was also found to perform better than only reassigning the contested points to the appropriate tree and breaking it accordingly.

\subsection{Homography and Localization}
Homography is a projective transformation between two perspective images of a planar scene that can also be used to determine the relative pose of an object with respect to a given reference image. The study of homography and localization of images and objects is textbook material; however, a brief overview is provided below as this process is used to experimentally test QuickMatch's performance. More information on both homography and object localization can be found in \cite{homography1} and \cite{homography2}. 

Given a the image coordinates $\widetilde{x}$ (expressed in homogeneous coordinates) of a point belonging to a planar surface as seen in a reference view, the image of the same point in a novel view can be found given the homography matrix $\overline{H}$ as $\overline{H}\widetilde{x}$ (again, expressed using homogeneous coordinates). The $\overline{H}$ matrix can be estimated with a set of known relative points (or matched features) between two views. To improve the estimate of $\overline{H}$, random sample consensus (RANSAC) is used to remove match outliers by randomly sampling the matches, finding a fit of the data, and then removing any matches that fall outside of a user defined region (see \cite{CVAlgo}).

\begin{figure*}[!b]
\centering
\captionsetup[subfigure]{justification=centering}
\subfloat[]{\includegraphics[width=.45\linewidth]{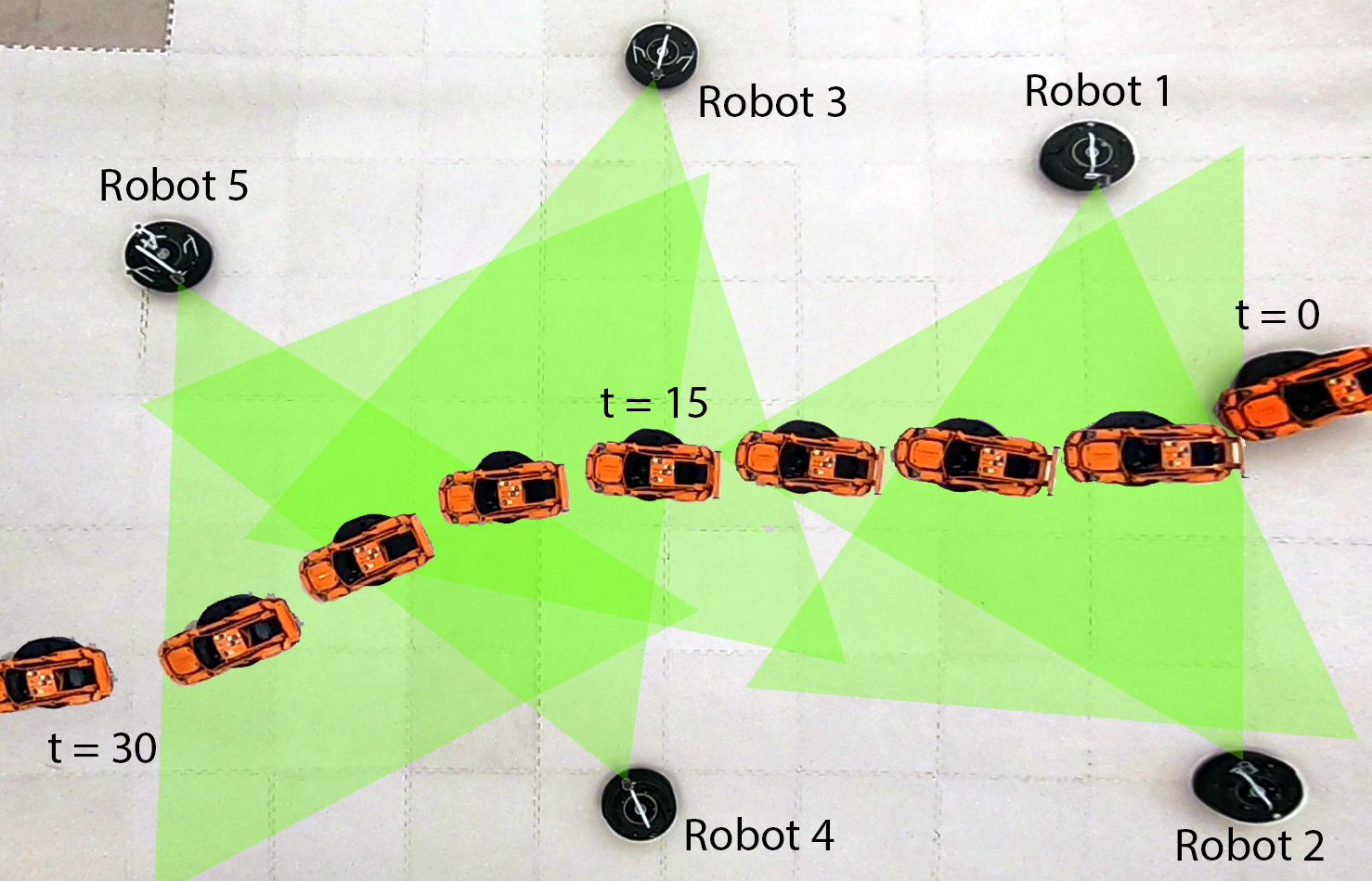}}
\hfill
  \subfloat[]{\includegraphics[width=.45\linewidth]{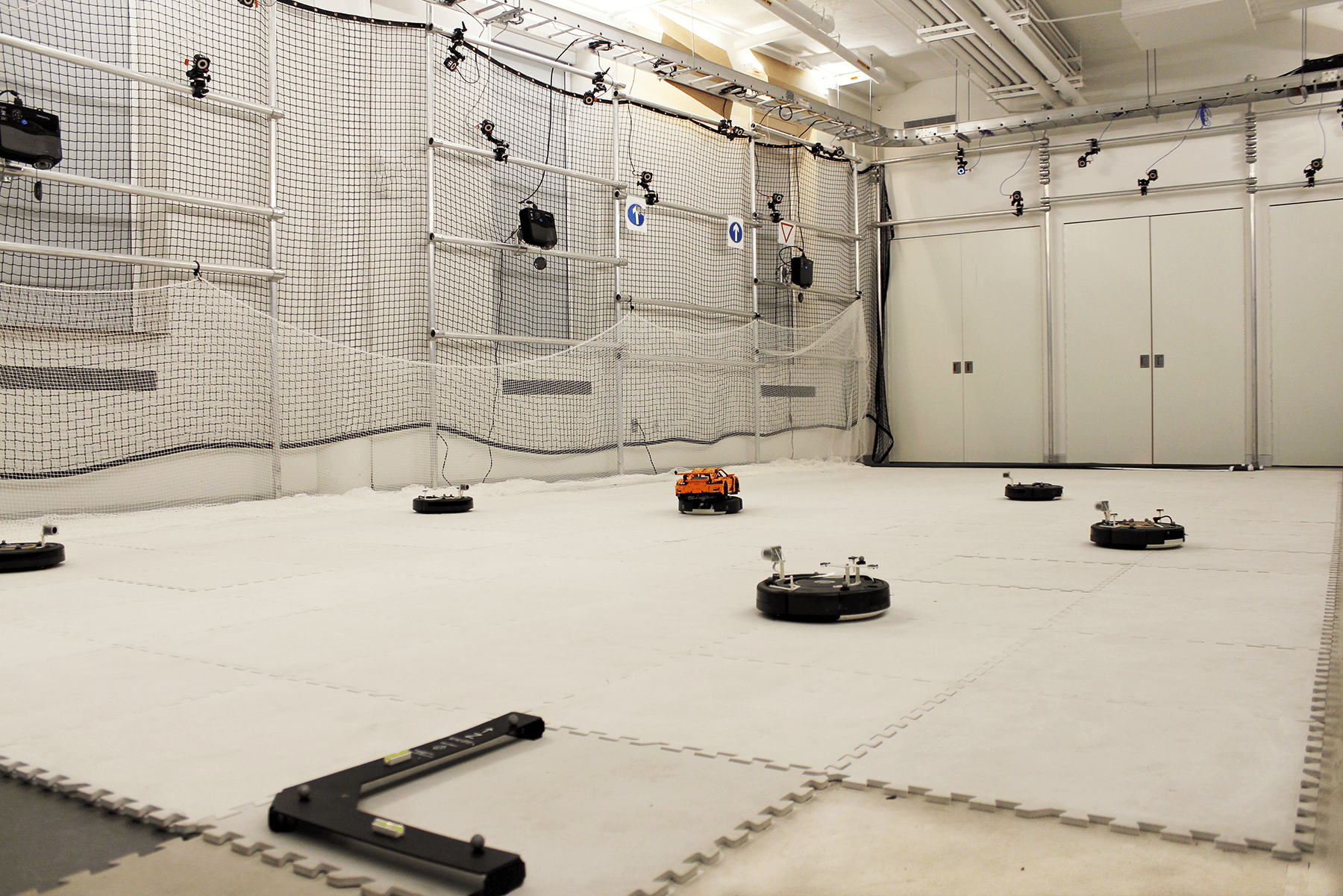}}
\caption{(a) Overhead view of experimental area with trajectory of the target object, position of the robots, and  the approximate field of view for the camera network (shown in yellow). (b) Prospective view of experimental area with modified iRobot Create2 platform, target object, and overhead OptiTrack\textsuperscript{\tiny{\textcopyright}} motion capture system.}
\label{field_robots}
\end{figure*}

Given $\overline{H}$, it is also possible to recover the relative pose between the two views; this can then be used to indirectly localize (recover the translation and rotation) of an approximately planar object in a relative coordinate system up to a distance scale factor, as shown in Figure \ref{field_robots} (a). Given some known size of the target object (e.g., height), the scale ambiguity can be resolved, recovering the full object relative position. In our applications, where each image is taken together with images from other cameras in the network, and where the pose of each camera is known, the target object can be accurately positioned in the global reference frame, allowing for the generation of a target's trajectory (e.g., Figure \ref{traj_figure}). 

Since localization using homography is limited to approximately planar surfaces, multiple reference images (as used here) are required to indentify different sides of an object.  Secondly, despite the use of robust estimation (RANSAC) inaccurate matchings are still possible, resulting in outlier measurements in distance and bearing. These inaccuracies are amplified by the sensitivity to object height estimate errors when calculating target distance. To account for these errors in practice, multiple measurements can be used to estimate each position, and then a filter can be used to smooth the target's trajectory (e.g. a Kalman Filter).

\section{Simulations}
This paper looks to compare primarily QuickMatch with the Distributed QuickMatch algorithm in simulation, and the QuickMatch algorithm with off the shelf matching tools in an experimental setting. For a detailed comparison of the performance of QuickMatch to other state-of-the-art matching methods (particularly multi-image matching methods), see \cite{tron_quickmatch}.

We begin with an illustrative example of Distributed QuickMatch on a synthetic 2 dimensional data set shown in Figures \ref{fig:sim1} and \ref{fig:sim2}. Figures \ref{fig:sim1} and \ref{fig:sim2} show a test case with 4 agents, and 250 feature points. There are 25 underlying clusters, generated with random Gaussian distributions around 25 evenly spaced points, each with 10 sample features. Figure \ref{fig:sim1} illustrates many of the steps in Algorithm \ref{algo:DQM}. Figure \ref{fig:sim1} (a) shows the $\ell^{-1}(a,0)$ labels for all of the agents. This mapping is created using Equations \ref{eq:kmeans1} and \ref{eq:kmeans2}, and the Voronoi partition seeds for the partition are shown as triangles. Note that the clusters along the boundaries have features of different colors, meaning these clusters would be split and improperly matched with just the naive partitioning approach. 

The partition generated by the Voronoi seeds can be seen in Figure \ref{fig:sim1} (b). Here, the $x_{min}$ point is plotted for each feature. It can be seen that at least three of the clusters are split by this partition. This is further shown in Figure \ref{fig:sim1} (c), which shows the result of QuickMatch being run on each agent partition individually. Most notably, the central cluster is split into three clusters. Each marker style represents a different cluster membership (i.e. points with the same marker belong to the same cluster).

Figure \ref{fig:sim1} (d) shows the clusters staged for initial agent switching. Each feature that is switched has a label of where it is being sent. Note that the central cluster has multiple labels, meaning even after the switch, the cluster will be segmented. Also, note that whole clusters are staged for transfer, and that the contested region is very conservative, since clusters even somewhat far from the boundary are being swapped. After the agents are switched, the agents check if the switched clusters are nearest to other contested points. Figure \ref{fig:sim1} (e) shows the reassignment of the clusters after this check is performed. Note that after the first swap, the points in the central cluster are all moving toward $a_0$. This highlights one drawback to this approach, if the boundaries are draw so that higher index agents have many contested clusters, the lower index clusters end up getting assigned to more features. 

\section{Experiment}
The experiment consists of a team of five iRobot Create2 ground robots, each with a forward facing camera, distributed throughout the experimental area shown in Figure \ref{field_robots}. Each camera has a $62 \degree \times 48 \degree$ field of view, and takes a $640 \times 480$ px image at 2 Hz. Through the center of the area, the target object is driven along the trajectory shown in Figure \ref{field_robots} (a) over approximately thirty seconds. All cameras are triggered simultaneously and the images are sent to a central computer for feature extraction and matching. The central computer has an Intel i7-7800x 3.5GHz processor, and runs Ubuntu 16.04 LTS and ROS Kinetic. Features are extracted using SIFT with an octave layer of 6, a contrast threshold of 0.10, an edge threshold of 15, and sigma of 1.0. The matches from QuickMatch (using $\rho = 1.1$) are used to determine which cameras observe the target object at each time step, based on the number of matches with a target image (in this experiment 10 matches are required). The matches between each reference images and the current images are used to determine the homography between them, using RANSAC with a threshold of 10.0. The homography is used to generate a bounding box around the target object using a perspective transformation on the target image corners. The relationship between pixel height of this box and distance from the camera is calibrated beforehand using an object of known size (in this experience a checkerboard pattern of know dimension). The localization points are recorded to build a target trajectory, which is then compared to ground truth measurements from an OptiTrack motion capture system (Figure \ref{field_robots} (b)).    

\begin{figure}[t]
  \centering
  \captionsetup[subfigure]{justification=centering}
  \subfloat[]{\includegraphics[width=.9\linewidth]{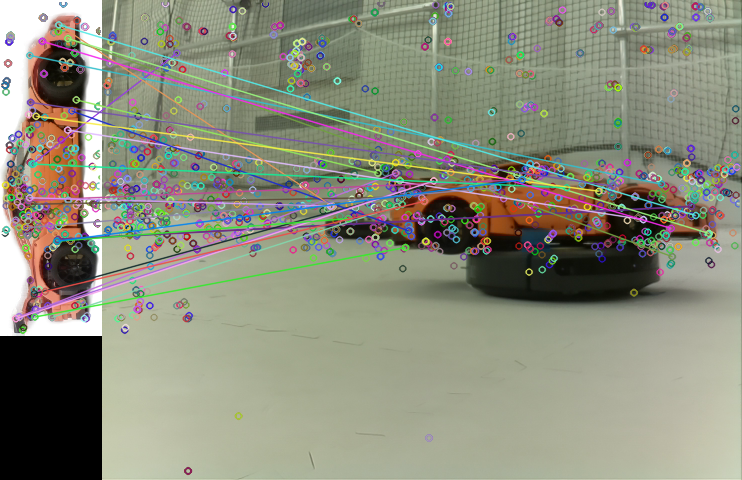}}\\
  \subfloat[]{\includegraphics[width=.9\linewidth]{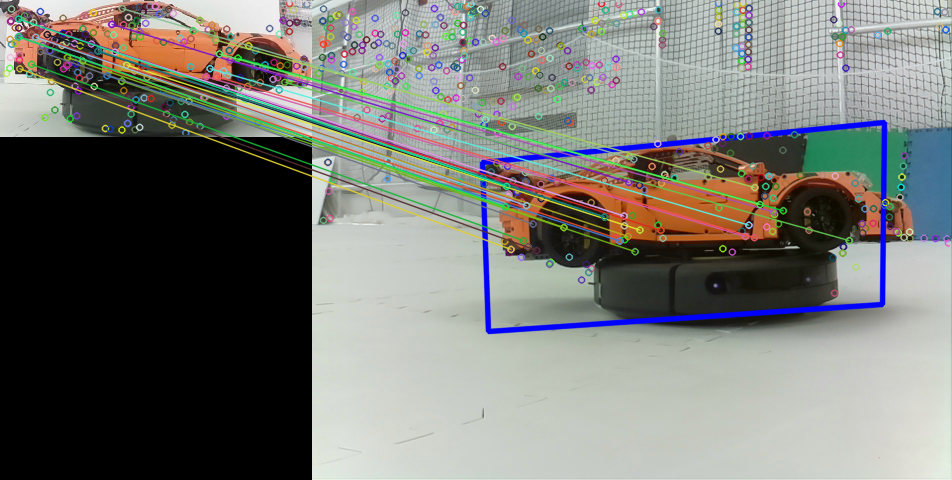}}
  \caption{(a) Example image matches between the reference object image (left) and an experimental image (right): Circles represent features, and lines indicate matches. (b) Homography and localization of car with prospective transform of bounding box.}
  \label{box_figure}
\end{figure}

\section{Results}
%
%\vspace{-5pt}
QuickMatch is evaluated in two ways: pure matching performance, and in the context of a target localization application. The QuickMatch algorithm is first compared to standard matching algorithms in the OpenCV Software Package  \pcite{opencv}, Brute Force (BF), and FLANN. Both algorithms use the Euclidean distance metric and a threshold match distance of 0.75 \pcite{opencv,SIFT}. Unlike QuickMatch, both algorithms cannot consider matches across more than two images but do have very low execution times. 

QuickMatch is implemented in Python and takes 5.6 seconds to find matches between 6254 SIFT features (from 115 images), while BF and FLANN are both implemented in C++, and both take approximately 0.05 seconds to find the matches between the reference image features, and the same 6254 features. This time difference arises from two factors: the inherently slower run time of Python compared to C++ \pcite{Code_speed}, and the extra comparisons done by QuickMatch to solve the entire Multi-match problem. If BF and FLANN compared all images with all other images combinatorially (as QuickMatch implicitly does) their computation times would be $\sim 5.75s$ seconds, which is comparable to QuickMatch's slower Python implementation. This time also does not account for the post processing time necessary to reconcile inconsistent matches from both BF and FLANN, is not required in QuickMatch.
%
%\vspace{-10pt}

\begin{figure}[t]
\centering
\subfloat{
  \includegraphics[width=.9\linewidth,trim={.5cm 0 .5cm .9cm},clip]{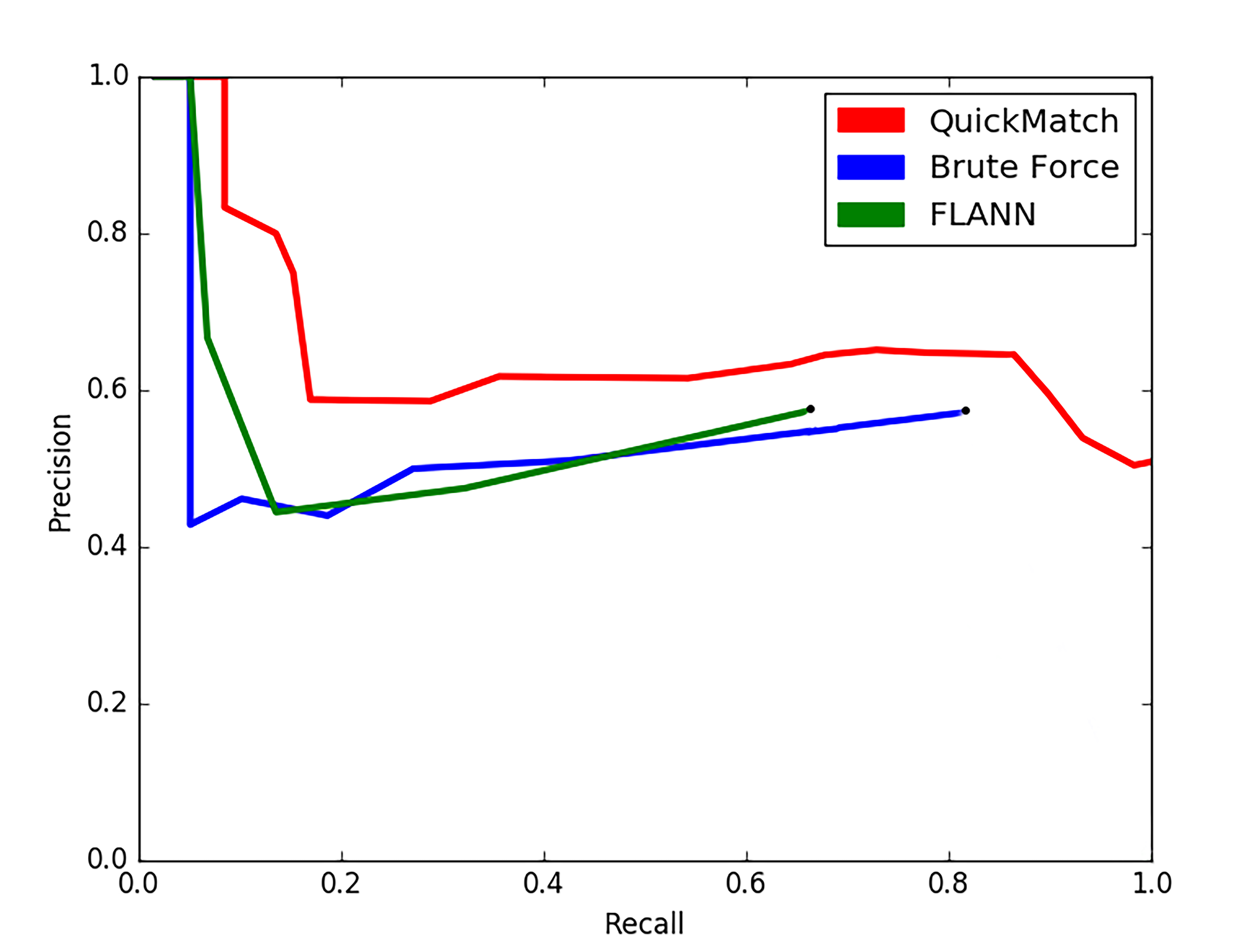}
  }
\caption{(a) Precision vs. recall curves for the QuickMatch, Brute Force, and FLANN algorithms. All algorithms are run on the same feature vectors. A match is considered to exist if the number of matched features is above a threshold. }
\label{rates_figure}
\end{figure}

\subsection{Precision Versus Recall}

Although QuickMatch is slower, it outperforms both BF and FLANN in the number of matches correctly found, and generally in terms of precision vs. recall (PR) and precision-recall area under the curve (PR AUC), which are common metrics for evaluating matching algorithms \cite{precision_recall}. Figure \ref{rates_figure} (a) shows the precision (fraction of correctly matched images) versus recall (fraction of possible matches found) curves for QuickMatch, BF, and FLANN. For any recall level, QuickMatch maintains a higher precision level than either BF or FLANN. Both BF and FLANN have terminations before a recall of 0.9 because at that level of discrimination, they are unable to find any matches in the data. QuickMatch on the other hand is still able to find some matches. These curves are non-monotonic because mismatched features appear at a higher rate than correctly matched features at higher thresholds. PR AUC is a threshold agnostic metric used for comparing overall performance of matching algorithms \pcite{precision_recall}. In terms of PR AUC, QuickMatch achieves 0.64, while BF and FLANN reach 0.49 and 0.45 respectively. The overall increase in precision stems for QuickMatch's ability to consider more instances of the reference object, by matching cycles of features across multiple images. It is therefore able to find the reference object not only more consistently, but with many more matched features. An example of these matches is shown in Figure \ref{box_figure}. 

\begin{figure*}[h]
\setlength{\fboxsep}{0pt}%
\setlength{\fboxrule}{0pt}%
\captionsetup[subfigure]{justification=centering}
\begin{center}
  \subfloat[]{
  \includegraphics[width=.46\linewidth]{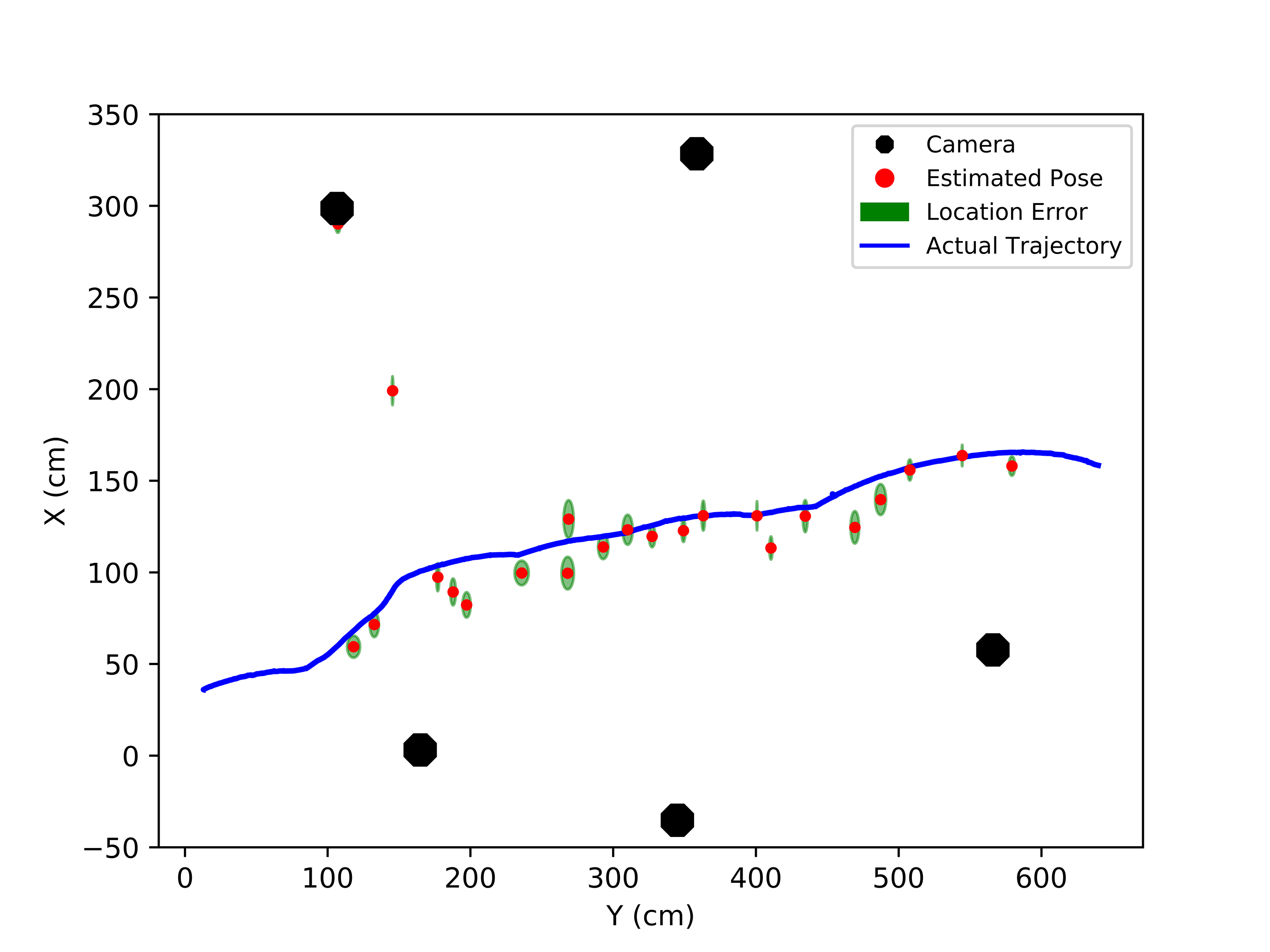}
  }
  \hfill
  \subfloat[]{
  \includegraphics[width=.46\linewidth]{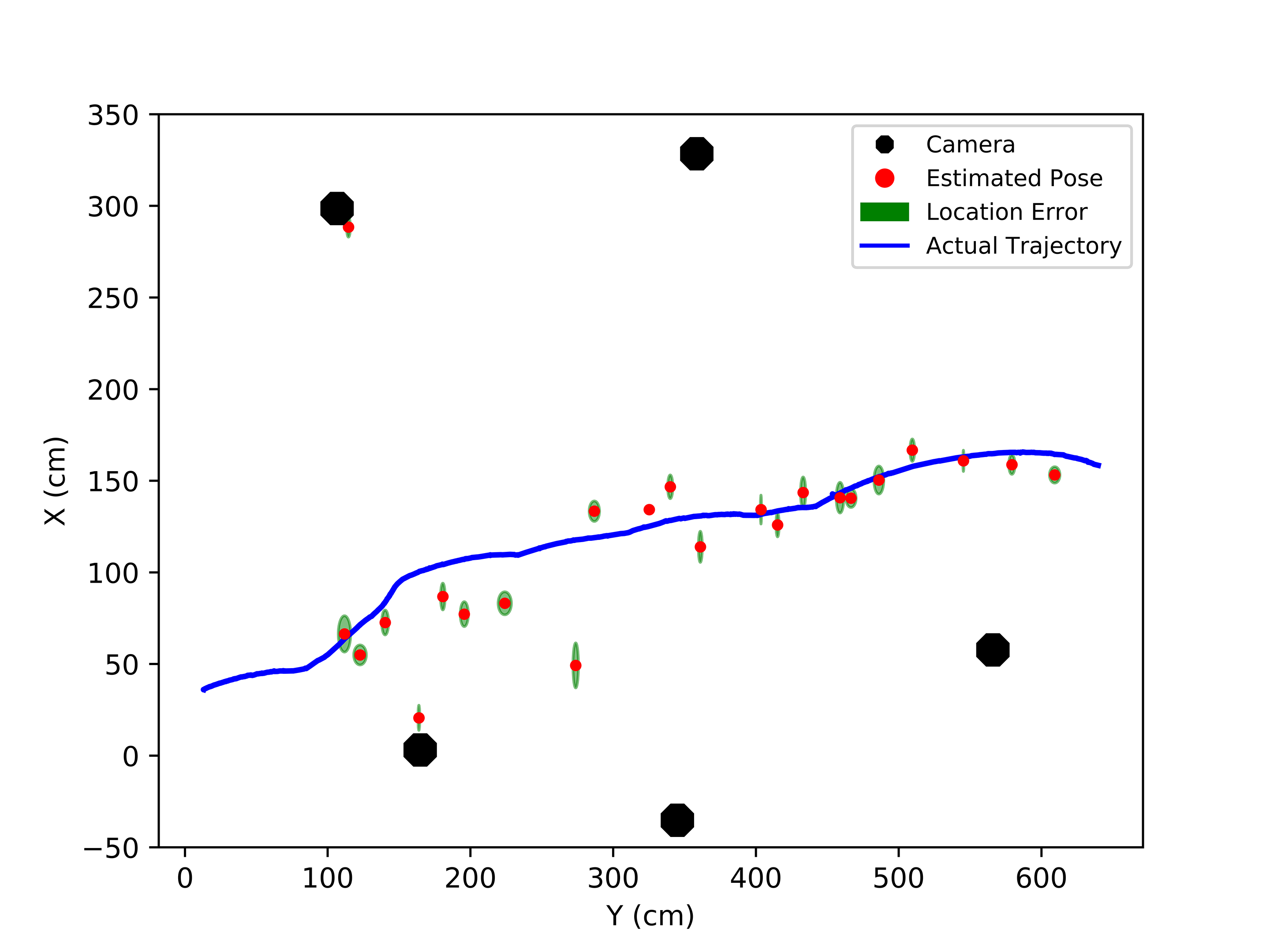}
  }
 \\
% \vspace{-10pt}
   %\hfill
  \subfloat[]{
  \includegraphics[width=.46\linewidth]{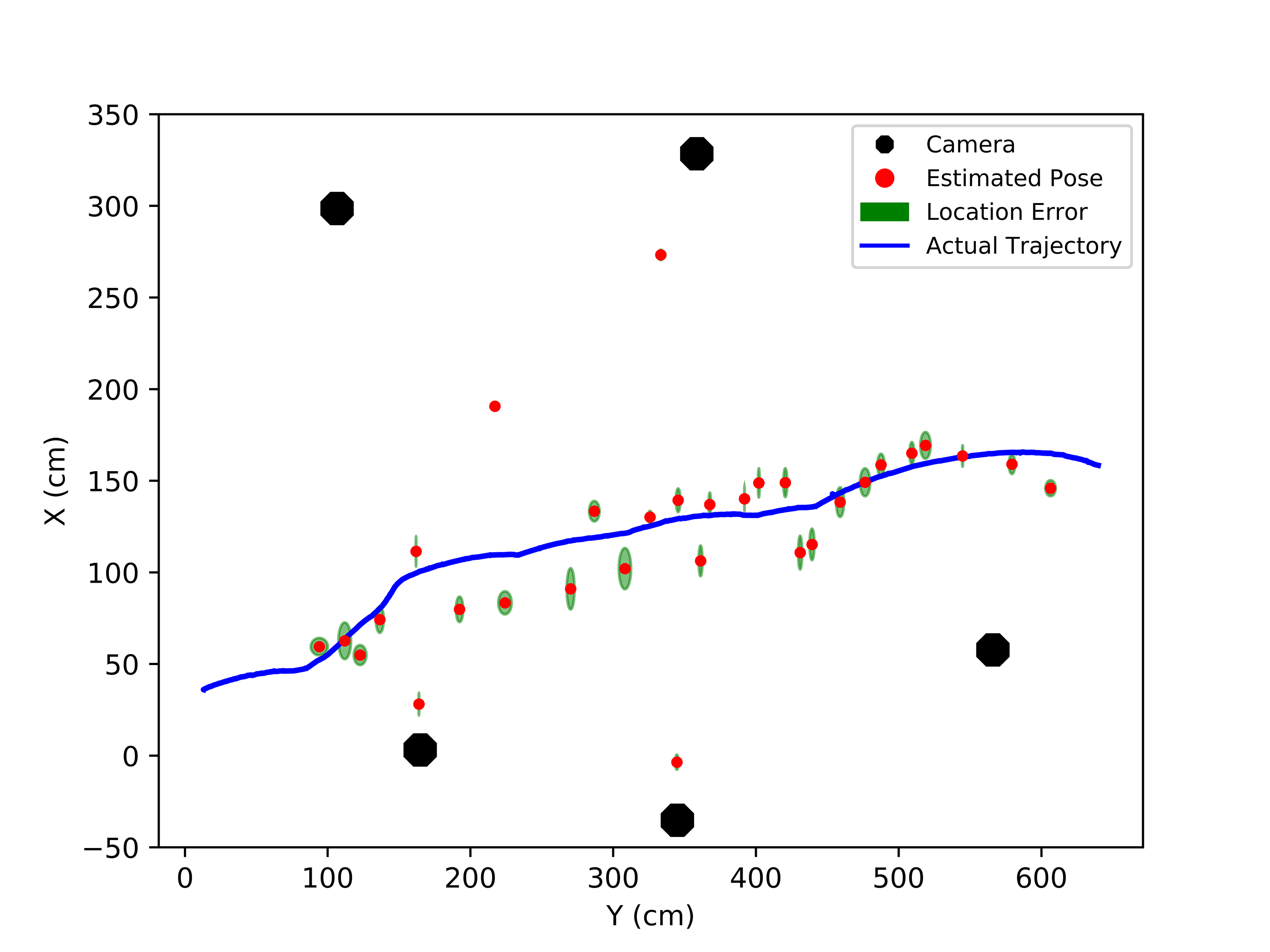}
  }
  \hfill
  \subfloat[]{
  \includegraphics[width=.46\linewidth]{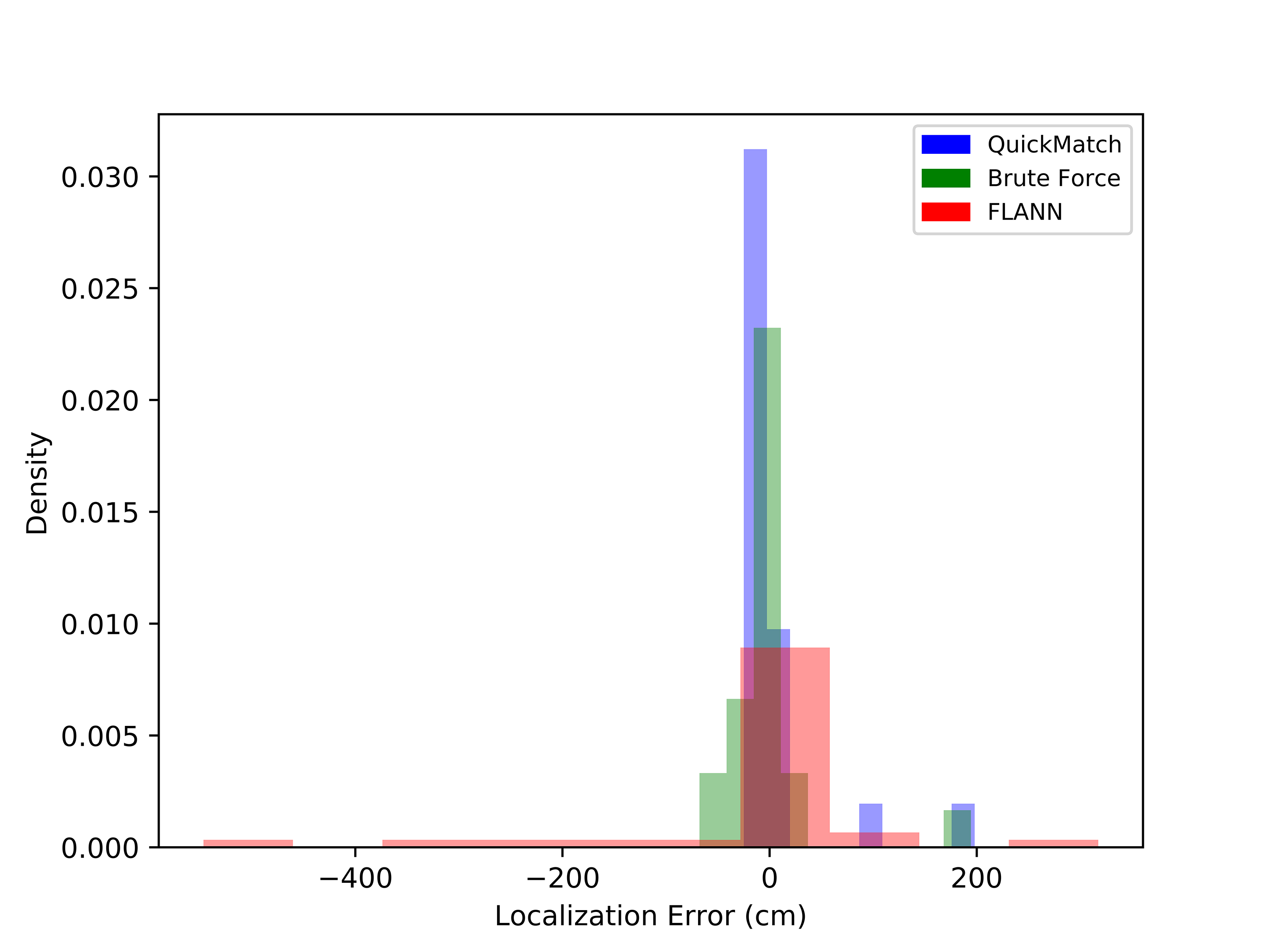}
  }
  \end{center}
\caption{(a) QuickMatch trajectory estimate. (b) BruteForce trajectory estimate. (c) FLANN trajectory estimate. (d) Histogram of estimate error for each algorithm.}
\label{traj_figure}
%\vspace{-5pt}
\end{figure*}
\subsection{Homography and Localization}
In order to further demonstrate the utility of the QuickMatch algorithm, matches were used to localize a target object in relation to the camera network, and then estimate its global trajectory. This was done using all three above algorithms with again an identical set of SIFT features. QuickMatch considers multi-image matches between the set of target images and the set of five robot images at each time step, while BF and FLANN consider matches between each target image and the robot image individually. Once feature matches are generated, RANSAC is used to estimate the homography matrix $\overline{H}$ for each pair of images while also removing outliers from the matches. The homography between the reference image and each robot image is used to generate a bounding box around the target in the robot image as shown in Figure \ref{box_figure} (b). This bounding box, given a known camera calibration, provides bearing  and height information for the target. The target height is known and is used to find the relate distance to the target with the bounding box height. With these two values, a distance and a bearing, the object can be localized with respect to each robot.

%\vspace{-4pt}

The above steps are performed using the match data from each of the three above algorithms. Figures \ref{traj_figure} (a-c) show the results of the localization estimation for each algorithm. Red points are estimate target poses for each time step, blue points denote the ground truth measurements, black octagons are the camera network positions, and the green regions are the one standard deviation error between all localization estimates at each time step. The localization error was found by taking the absolute distance between the estimated and ground truth position at each time step. QuickMatch had an error of $0.2118 \pm 0.4254$ meters, BF had an error of $0.2349 \pm 0.4027$ meters, and FLANN had an error of $0.6232 \pm 1.1722$ meters. QuickMatch outperforms both BF and FLANN in terms of accuracy, which is indicative of its higher match quality. BF matcher also performs well and maintains a low variance, however it is not as accurate. FLANN is the worst performing of the three, and has a number of extremely erroneous estimates. Generally, monocular camera distance measurements are very sensitive to match errors, meaning target localization error is an indirect method for testing the overall accuracy of each method. Figure \ref{traj_figure} (d) shows a histogram of the localization error, which is found by comparing the localization estimate to the ground truth pose at each time step. The histogram makes it clear that QuickMatch maintains a higher number of accurate matches and has a small number of highly erroneous estimates. In practical applications, a Kalman filter would be employed to smooth the estimates, but the values are left unaltered here to demonstrate the algorithm's output.

\begin{table*}[b]
\caption{Comparison of QuickMatch to Distributed QuickMatch by agent number on graffiti data set.}
\resizebox{\textwidth}{!}{%
\begin{tabular}{|c|c|c|c|c|c|c|c|c|c|c|c|c|c|}
\hline
Number of Agents & 1 & 2 & 3 & 4 & 5 & 6 & 7 & 8 & 9 & 10 & 15 & 20 & 25 \\ \hline
\begin{tabular}[c]{@{}c@{}}Compute Time \\ Per Agent (s)\end{tabular} & 1.37 & 21.53 & 24.62 & 25.64 & 27.72 & 27.93 & 29.40 & 29.06 & 29.29 & 30.00 & 31.21 & 30.59 & 30.68 \\ \hline
\begin{tabular}[c]{@{}c@{}}Post-QP Compute Time \\ Per Agent (s)\end{tabular} & \multicolumn{1}{l|}{1.37} & \multicolumn{1}{l|}{5.14} & \multicolumn{1}{l|}{3.81} & \multicolumn{1}{l|}{2.25} & \multicolumn{1}{l|}{2.53} & \multicolumn{1}{l|}{1.95} & \multicolumn{1}{l|}{1.85} & \multicolumn{1}{l|}{1.05} & \multicolumn{1}{l|}{1.12} & \multicolumn{1}{l|}{1.18} & \multicolumn{1}{l|}{0.77} & \multicolumn{1}{l|}{0.72} & \multicolumn{1}{l|}{0.67} \\ \hline
\begin{tabular}[c]{@{}c@{}}QP Time\\  Per Agent (s)\end{tabular} & NA & 16.39 & 20.81 & 23.39 & 25.18 & 25.98 & 27.53 & 27.46 & 27.76 & 28.59 & 30.08 & 29.82 & 30.00 \\ \hline
\begin{tabular}[c]{@{}c@{}}Percent Contested\\ Clusters\end{tabular} & 0 & 18.21 & 28.62 & 27.51 & 21.43 & 29.24 & 35.42 & 46.34 & 40.13 & 39.78 & 42.87 & 48.71 & 52.16 \\ \hline
\begin{tabular}[c]{@{}c@{}}Number of\\ Clusters Found\end{tabular} & 1320 & 1313 & 1314 & 1278 & 1265 & 1264 & 1281 & 1258 & 1269 & 1264 & 1246 & 1259 & 1309 \\ \hline
\begin{tabular}[c]{@{}c@{}}\% Contested Features \\ Found\end{tabular} & NA & 99.80 & 99.87 & 99.87 & 99.83 & 99.64 & 99.90 & 99.77 & 99.91 & 99.63 & 99.57 & 98.24 & 99.1 \\ \hline
\end{tabular}%
}
\label{table:result}
\end{table*}
\begin{figure*}[!b]
\label{fig:sim2}
\centering
\captionsetup[subfigure]{justification=centering}
  \subfloat[]{
  \includegraphics[width=.48\linewidth]{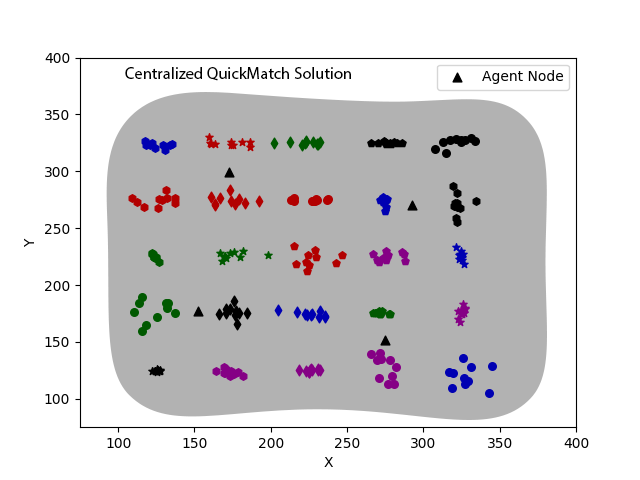}
  }
  \subfloat[]{
  \includegraphics[width=.48\linewidth]{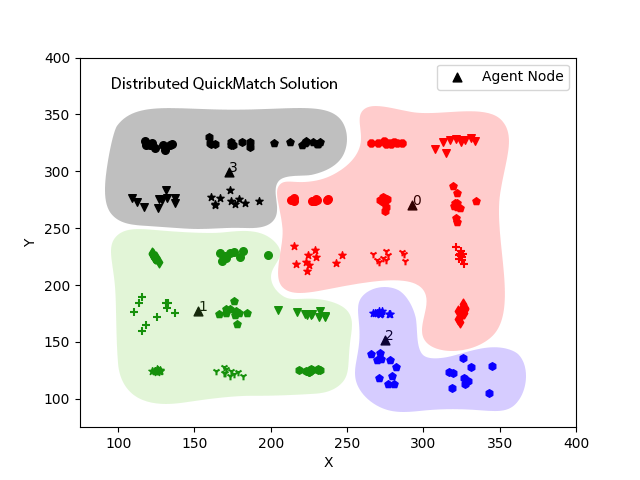}
  }
  \caption{(a) Clustering result from the centralized QuickMatch algorithm on the synthetic data set. Feature color denotes cluster membership. (b) Clustering result from the Distributed QuickMatch algorithm on the synthetic data set. Feature color denotes agent membership.}
  \label{fig:sim2}
\end{figure*}

\subsection{Centralized vs Distributed Comparison}

To test the difference between the distributed and centralized approaches, we look at how many clusters are split by naive partitions in the feature space to determine if the distribution scheme above is even warranted. This test is performed on images from the Graffiti data set (http://www.robots.ox.ac.uk/˜vgg/data/data-aff.htm), in order to simulate realistic conditions where the ground truth is unknown. Given the clusters of features determined by QuickMatch, $C_c \in \mathcal{M}$, we define the count of each agent membership in the cluster as \begin{equation}
    q_a(C_c) = \vert C_c \cap \ell^{-1}(a) \vert,
\end{equation}
where $q_a$ is the number of features in $C_c$ with label $a$. With this we can define the split quality $Q$ of a cluster $C_c$ as 
\begin{equation}
    Q(C_c) = \frac{\max\limits_{a \in A}(q_a)}{\vert C_c \vert}.
\end{equation}
With this quality metric, we can quantify the number of contested clusters in a given partition as $C_c:Q(C_c)<1$ and the percent of contested clusters as
\begin{equation}
    p_{\textrm{contested}}=\frac{\lvert C_c\in\mathcal{M}: Q(C_c)<1\rvert}{\lvert C_c\rvert}.
\end{equation}
With this metric $p_{\textrm{contested}}$, we evaluate both two methods for creating the initial seeds for the Voronoi partition, as well as the ability for Distributed QuickMatch to find and appropriately reassign contested clusters. For the synthetic data set, the ground truth is known, however for the graffiti data set, we assume that the QuickMatch clusters are the ground truth.

To determine how many contested features, and hence split clusters, are missed by Distributed QuickMatch, we calculate the following precision
\begin{equation} 
     p_{\textrm{split}}=\frac{\lvert S_A \rvert}{\sum_{C_c:Q(C_c)<1}\lvert C_c\rvert}
\end{equation} 
where $S_A$ is the set of all contested features in $K_\mathcal{I}$. In other words, we can consider Distributed QuickMatch to be in part as a classifier that needs to detect which features are contested; then, $p_{\textrm{split}}$ represents the precision of such classifier (number of features that are declared as contested over the number of features that ought to be declared). 

\begin{figure*}[t]
  \centering
  \captionsetup[subfigure]{justification=centering}
  \subfloat[]{\includegraphics[width=.75\linewidth]{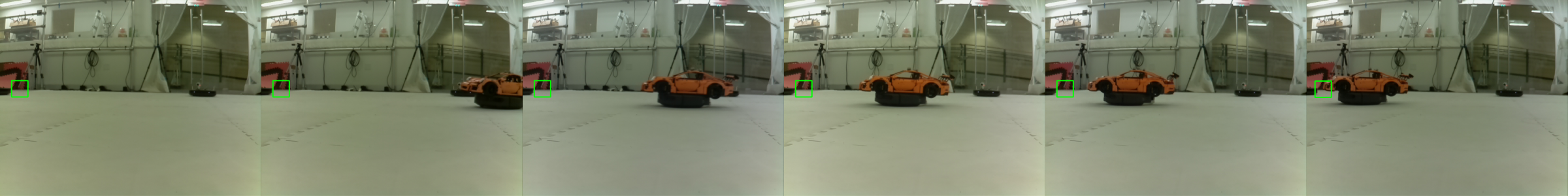}}\\
  \subfloat[]{\includegraphics[width=.75\linewidth]{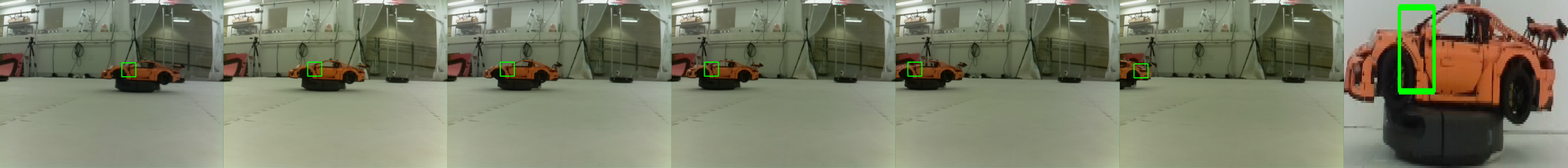}}
  \caption{(a) Landmark feature cluster. (b) Target feature cluster.}
  \label{disc_figure}
\end{figure*}

The results of these tests are shown in Table \ref{table:result}. The first column in Table \ref{table:result} shows the results of the centralized QuickMatch algorithm. Row five shows that as the number of agents increases, intuitively, the percentage of contested clusters also increases. At the same time however, the post-QP computation time decreases as agent number decreases, because each agent has to work on considerably fewer features. The largest computational requirement in the Distributed QuickMatch algorithm is finding the boundary distances with the QP. The time reported for this is per-agent, however it is worth noting the implementation of this QP is sub-optimal. In the QP formulation, we consider each partition individually, meaning we solve $m$ QPs for each feature. In the future, we plan to reduce this to combining all of the constraints to solve a single QP and this is a focus of future work. One key aspect of Distributed QuickMatch to note is that is finds around 99 percent of all contested points, meaning it is very good at finding split clusters. One interesting result of the Distributed QuickMatch algorithm is that it matches the features into fewer, larger clusters due to the use of its finite density kernel. This is a counter-intuitive result and will be a subject of future study for this algorithm. 

The final clustering results from the synthetic data set in the Simulation section are shown in Figure \ref{fig:sim2}. Figure \ref{fig:sim2}(a) shows the result of the centralized QuickMatch algorithm run on the synthetic data, while Figure \ref{fig:sim2}(b) shows the final result from Distributed QuickMatch. Note that many of the features from the higher index agents have been shifted to the lower index agents, but ultimately each cluster has features belonging to only one agent. Ultimately both algorithms are able to cluster all of the features correctly.

\subsection{Feature Discovery}\label{sec:featurediscovery}
The QuickMatch algorithm implicitly discovers common features among images by creating clusters of similar features. These clusters correspond to specific locations in the universe, and therefore can be used to find both targets and landmarks across images. Landmarks, although not used in this paper, are points that occur commonly across all images (except when occluded), and are useful for multi-agent localization tasks. In the experimental images collected, landmarks were the clusters with the largest number of features, because many of this images did not contain the target object. An example landmark cluster is shown in Figure \ref{disc_figure} (a). Features belonging to the target object are generally smaller than the landmark clusters, but can still be extracted, and show key features of the target. Figure \ref{disc_figure} (b) shows one such cluster, which is the front hood of the car model. Feature discovery is one attribute of QuickMatch that does not exist in either BF or FLANN and can be useful for discerning what features are most descriptive of images from the network.

\section{Conclusion}
This paper highlights the utility of QuickMatch multi-image matching for object matching and presents the Distributed QuickMatch algorithm. QuickMatch is able to find many more object feature matches than standard methods by considering matches across all images, not just pairwise matches. The presented experiment tests the QuickMatch algorithm in an experimental setting with realistic conditions, and shows that multi-image matching is superior to standard methods at matching the reference object (even as it enters and exits images across the entire camera network). QuickMatch is also tested with a target object localization and again outperforms both the BF and FLANN algorithms. Beyond testing QuickMatch, we demonstrate the Distributed QuickMatch algorithm on both the graffiti and synthetic data sets. We also demonstrate QuickMatch's feature discovery ability by showing a characteristic landmark and target feature cluster from the test images. This approach is the precursor to an online and decentralized approach. Our future work will focus on the online version of object discovery and localization and multi-camera homography. We also plan to decrease the Distributed QuickMatch's QP implementation computation time. Overall, QuickMatch is shown to be a versatile multi-feature matching algorithm that outperforms standard pairwise matching algorithms, and Distributed QuickMatch offers an avenue for the QuickMatch framework to handle a large volume of features with minimal inter-agent communication. 

\begin{acks}
This work was supported by the National Science Foundation under grants NRI-1734454, and IIS-1717656.
\end{acks}

%
% ---- Bibliography ----
%

%Citations go in alphabetical order

\begin{comment}

\end{comment}

\end{document}